\def\year{2018}\relax
\documentclass[letterpaper]{article} 
\usepackage{aaai18}  
\usepackage{times}  
\usepackage{helvet}  
\usepackage{courier}  
\usepackage{url}  
\usepackage{graphicx}  
\usepackage{amsfonts} 
\begin{document}
%
\title{Guiding Reinforcement Learning Exploration Using Natural Language}
\author{Brent Harrison \\ Department of Computer Science \\ University of Kentucky \\ Lexington, Kentucky, USA \\ harrison@cs.uky.edu \And Upol Ehsan \\ School of Interactive Computing \\ Georgia Institute of Technology \\ Atlanta, Georgia, USA \\ ehsan@gatech.edu \And Mark O. Riedl \\ School of Interactive Computing \\ Georgia Institute of Technology \\ Atlanta, Georgia, USA \\ riedl@cc.gatech.edu
}

\nocopyright

\maketitle
\begin{abstract} 
In this work we present a technique to use natural language to help reinforcement learning generalize to unseen environments. 
%
%
This technique uses neural machine translation, specifically the use of encoder-decoder networks, to learn associations between natural language behavior descriptions and state-action information. 
We then use this learned model to guide agent exploration using a modified version of policy shaping to make it more effective at learning in unseen environments. 
We evaluate this technique using the popular arcade game, Frogger, under ideal and non-ideal conditions. 
This evaluation shows that our modified policy shaping algorithm improves over a Q-learning agent as well as a baseline version of policy shaping. 
\end{abstract} 

\section{Introduction}
Interactive machine learning (IML) algorithms seek to augment machine learning with human knowledge in order to enable intelligent systems to better make decision in complex environments. 
These algorithms allow human teachers to directly interact with machine learning algorithms to train them to learn tasks faster than they would be able to on their own. 
Typically, humans interact with these systems by either providing demonstrations of positive behavior that an intelligent agent can learn from, or by providing online critique of an agent while it explores its environment. 
While these techniques have proven to be effective, it can sometimes be difficult for trainers to provide the required demonstrations or critique. 
Demonstrations may require that the trainer possess in-depth prior knowledge about a system or its environment, and trainers may have to provide hundreds of instances of feedback before the agent begins to utilize it.
This issue is compounded when one considers that this training must occur for each new environment that the agent finds itself in. 

In this work, we seek to reduce the burden on human trainers by using natural language to enable interactive machine learning algorithms to better generalize to unseen environments. 
Since language is one of the primary ways that humans communicate, using language to train intelligent agents should come more naturally to human teachers than using demonstrations or critique.
In our proposed approach, natural language instruction also need not be given online while the agent is learning. 
Allowing instruction to be given offline greatly reduces the time and effort required on the part of the human teacher to train these intelligent agents.

Humans are also extremely proficient and generalizing over many states, and often language aids in this endeavor. 
With this work, we aim to use human language to learn these human-like state abstractions and use them to enhance reinforcement learning in unseen environments. 
To do that, we use neural machine translation techniques---specifically encoder-decoder networks---to learn generalized associations between natural language behavior descriptions and state/action information. 
We then use this model, which can be thought of as a model of generalized action advice, to augment a state of the art interactive machine learning algorithm to make it more effective in unseen environments. 
For this work, we choose to modify \textit{policy shaping}, an interactive machine learning algorithm that learns from human critique~\cite{griffith2013policy}.

We evaluate this technique using the arcade game, Frogger. 
Specifically, we evaluate how our technique performs against a base Q-learning algorithm and a version of policy shaping that uses only demonstrations as examples policy critique on the task of learning on a set of unseen Frogger maps in a variety of situations

To summarize, the main contributions of this paper are as follows: 1) We show how neural machine translation can be used to create a generalized model of action advice, 2) we show how this model can be used to augment policy shaping to enable reinforcement learning agents to better learn in unseen environments, and 3) we perform an evaluation of our method in the arcade game, Frogger, on several previously unseen maps using unreliable synthetic oracles meant to simulate human trainers. 


\section{Related Work} 
\label{sec:related-work}
This work is primarily related to two bodies of artificial intelligence research: interactive machine learning and knowledge transfer in reinforcement learning. 
Interactive machine learning (IML)~\cite{chernova2014robot} algorithms use knowledge provided by human teachers to help train machine learning models. 
This allows for human experts to help train intelligent agents, thus enabling these agents to learn faster than they would if left to learn on their own.
Typically human teachers interact with the agent by providing either demonstrations of correct behavior~\cite{argall2009survey} or directly critique the agent's behavior~\cite{loftin2014learning,griffith2013policy,cederborg2015policy}.
Our work seeks to improve upon these methods by enabling these algorithms to learn from natural language in addition to demonstrations or critique. 

There has been other work on using natural language to augment machine learning algorithms. 
There has been much work done on using natural language instructions to help reinforcement learning agents complete tasks more efficiently.
Early works in this area focused on learning mappings between these instructions and specific control sequences in learning environments~\cite{branavan2009reinforcement,branavan2010reading,matuszek2013learning}.
In this previous work, language is used mainly used to instruct how to complete a specific task in a specific environment. 
In other words, language and state are tightly coupled. 
The main way that our work differs from this work is that we are seeking to use language as an abstraction tool. 
Our work focuses on exploring how language can be used to help reinforcement learning agents transfer knowledge to unseen environments. 

More recent work has examined how language can help reinforcement learning agents in a more environment-agnostic way. 
For example, work has been done on using high-level task specifications to engineer environment-agnostic reward functions to improve learning~\cite{macglashan2015grounding}. 
Also, techniques such as sentiment analysis have been used to bias agent exploration to improve learning in unseen environments~\cite{krening2017learning}.
Most of these techniques, however, require additional information about the environment, such as descriptions of object types in the environment, that may not always be readily available.
Our technique relaxes this requirement by using neural machine translation to learn relationships between natural language action/state descriptions and parts of the state space. 

The work most closely related to our own involves using deep Q-learning to identify language representations that can help reinforcement learning agents learn in unseen environments~\cite{narasimhan2015language}. 
This technique, however, also requires some knowledge about the environment to be provided in order to learn these representations. 
Our technique does not require additional information to be provided by the domain author as all state annotations are generated by human teachers. 

\section{Background}
\label{sec:background}
In this section, we will discuss three concepts that are critical to our work:reinforcement learning, policy shaping, and encoder-decoder networks. 
\subsection{Reinforcement Learning}
Reinforcement learning~\cite{suttonbarto1998} is a technique that is used to solve a Markov decision process (MDP).
A MDP is a tuple $M = <S,A,T,R,\gamma>$ where $S$ is the set of possible world states, $A$ is the set of possible actions, $T$ is a transition function $T: S \times A \rightarrow P(S)$, $R$ is the reward function $R : S \times A \rightarrow \mathbb{R}$, and $\gamma$ is a discount factor $0 \leq \gamma \leq 1$.

Reinforcement learning first learns a policy $\pi : S \rightarrow A$, which defines which actions should be taken in each state. 
In this work, we use Q-learning~\cite{watkins1992}, which uses a Q-value $Q(s,a)$ to estimate the expected future discounted rewards for taking action $a$ in state $s$.
As an agent explores its environment, this Q-value is updated to take into account the reward that the agent receives in each state. 
In this paper, we use Boltzmann exploration~\cite{watkins1989models} to select the actions that a reinforcement learning agent will take during training.
When using Boltzmann exploration, the probability of the agent choosing a particular action during training is calculated as $Pr_{q}(a)= \frac{e^{Q(s,a)/\tau}}{\sum_{a'}e^{Q(s,a')/\tau} }$, where $\tau$ is a temperature constant that controls whether the agent will prefer more random exploration or exploration based on current Q-values. 

\subsection{Policy Shaping}
In this paper, we build upon the policy shaping framework~\cite{griffith2013policy}, which is a technique that incorporates human critique into reinforcement learning. 
Unlike other techniques such as reward shaping, policy shaping considers critique to be a signal that evaluates whether the \textit{action} taken in a state was desirable rather than whether the resulting \textit{state} was desirable. 
Policy shaping utilizes human feedback by maintaining a \textit{critique policy} to calculate the probability, $Pr_{c}(a)$, that an action $a \in A$ should be taken in a given state according to the human feedback signal. 
During learning, the probability that an agent takes an action is calculated by combining both $Pr_{c}(a)$ and $Pr_{q}(a)$:
\begin{equation}
Pr(a) = \frac{Pr_{q}(a)Pr_{c}(a)}{\sum_{a'\in A}Pr_{q}(a')Pr_{c}(a')}
\end{equation}

Thus, the ultimate decision on which action to explore during learning is a combination of knowledge from the agent's experience as well as knowledge from a human teacher. 

The critique policy used in policy shaping is generated by examining how consistent the feedback for certain actions are. 
If an action receives primarily positive or negative critique, then the critique policy will reflect this with a greater or lower probability, respectively, to explore that action during learning. 
\begin{figure*}
\centering
\includegraphics[scale=0.60]{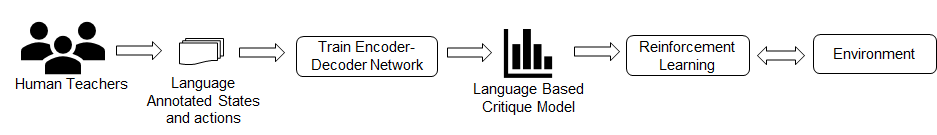}
\caption{High-level flowchart of our technique.}
\label{fig:flowchart}
\end{figure*}
\subsection{Encoder-Decoder Networks}
Encoder-decoder networks have been used frequently in other areas, such as machine translation~\cite{luong-pham-manning:2015:EMNLP}, to learn how to convert sets of input sequences into desired output sequences. 
In this work we use encoder-decoder networks to translate state/action descriptions written in natural language into machine-understandable state/action information that the natural language describes.
For example, the input to this network could be natural language describing the layout of a grid environment and an action taken in that specific state, while the desired output of the network would be the specific state and action representation used by the learning agent. 

This translation task, sometimes known as sequence-to-sequence learning, involves training two recurrent neural networks (RNNs): an encoder network and a decoder network.
In this generative architecture, these component neural networks work in conjunction to  learn how to translate an input sequence $X = (x_1, ..., x_T)$ into an output sequence $Y = (y_1, ..., y_T')$. 
To do this, the encoder network first learns to encode the input vector $X$ into a fixed length context vector $v$.
This context vector is meant to encode important aspects of the input sequence to aid the decoder in producing the desired output sequence.
This vector is then used as input into the second component network, the decoder, which is a RNN that learns how to iteratively decode this vector into the target output $Y$.
By setting up the learning problem in this particular way, this thought vector encodes high-level concept information that can help the decoder construct general state representations for each input sequence.

\section{Using Language to Generalize Human Critique}
\label{sec:methods}
As mentioned previously, one of the primary disadvantages of interactive machine learning is that humans must retrain the agent whenever it encounters a new environment. 
To address this issue, we show how an encoder-decoder network~\cite{luong-pham-manning:2015:EMNLP} can be used to learn a \textit{language-based critique policy}. 
A high level overview of our technique can be seen in Figure~\ref{fig:flowchart}.
Our technique works by first having humans generate a set of annotated states and actions by interacting with a single learning environment offline and thinking aloud about the actions they are performing.
These annotations are then used to train an encoder-decoder network to create the language-based critique model. 
This model can then be queried while the agent is exploring new environments to receive action advice to guide it towards states with potentially high rewards. 
This can be done even if the learning agent encounters states that have not been explicitly seen before and have not been used to train the language-based critique model. 
Each of these steps will be discussed in greater detail below. 


\subsection{Acquiring Human Feedback}
Typically, training an agent using critique requires a large amount of consistent online feedback in order to build up a \textit{critique policy}, a model of human feedback for a specific problem.
In other words, a human trainer would normally be required to watch an agent as its learning and provide feedback which is used in real time to improve the agent's performance.
This is because critique normally comes in the form of a discrete positive or negative feedback signal that is then associated with a given state or action. 
This provides the agent with little opportunity to generalize to unseen environments since this feedback is tightly coupled with state information. 
To address this, our technique uses natural language as a means to generalize feedback across many, possibly unknown, states.
We do this by training an encoder-decoder model to act as a more general critique policy that we refer to as a \textit{language-based critique policy}, which enables an agent to receive action advice for any potential state it finds itself in. 

We use two types of data in order to create this policy: examples of actions taken in the environment and natural language describing the action taken. 
This information can be gathered by having humans interact with the agent's learning environment while providing natural language descriptions of their behavior.
There are many ways that humans can potentially provide these state and action annotations. 
For instance, humans could provide a full episode of behavior along with behavior annotations. 
It is also possible for humans to provide incomplete trajectories, or even simply examples of single actions, along with natural language annotations.
Regardless of how they were collected, the state/action demonstrations provided by the human can be stored and later used as a positive feedback signal while the language can be used to help generalize this feedback signal over many states. 

For an example of this process, consider an environment where an agent is tasked with dodging obstacles. 
Assume in this environment that the learning agent can only move in the four cardinal direction: up, down, left, and right. 
In order to gather the necessary data to learn the feedback policy, a human trainer is presented with different obstacle initializations and tasked with providing short behavior examples of navigating them. 
The trainer could then provide feedback on each action that they took after the fact. 
An example of this feedback might be the description, \textit{I am dodging the obstacle that is coming up beside me,} if describing the action to move up when an obstacle is approaching from the side of the agent. 


\subsection{Training the Encoder-Decoder Network}
Due to how these annotations are generated, it is possible to directly associate language information to state and action information. 
This paired data is used to train an encoder-decoder network. 
Specifically, the natural language descriptions are used as inputs to the network, and the network is then tasked with reconstructing the state and action that are associated with that description. 
By asking the network to reconstruct the state and action, the network learns to identify common elements shared between similar inputs and, importantly, how these elements in the input natural language sequences relate to certain regions of the output sequence of state and action information.
This enables the network to learn high-level concept information that enables it to generalize natural language advice to unseen states.

\subsection{Utilizing the Language-Based Critique Policy}
The ultimate goal of this work is to use the language-based critique policy to speed up learning in unseen environments. 
To do that, we use this language-based critique policy in conjunction with policy shaping. 
Recall that a reinforcement learning agent using policy shaping makes decisions using two distinct pieces of information: $Pr_{q}(a)$, the probability of performing an action based on that action's current Q-value, and $Pr_{c}(a)$, the probability of performing an action based on the human critique policy. 
Here, we will use the language-based critique policy learned early to take the place of the standard critique policy normally used by policy shaping.  

For the language-based critique policy to be used in this framework, we must be able to calculate the probability of performing an action in each state, even if it has never been seen before. 
Whenever the agent encounters a state, we can query the language-based critique policy to get the probability of performing each action in that state for a given natural language input; however, for this to be of any use we must first determine which piece of feedback out of all of the feedback used in the training set is most applicable to the current state.
For each natural language utterance in our training set and for each action the agent perform in its current state, we calculate the log probability of the model reconstructing the agent's current state and performing said action. d
One can think of this log probability as how well an utterance describes performing that specific action in that specific state. 
Whichever utterance is associated with the action that results in the overall largest log probability is then used as the network input to create the action distribution.
To create the action distribution, we calculate the following:

\begin{figure*}
\centering
\includegraphics[scale=0.7]{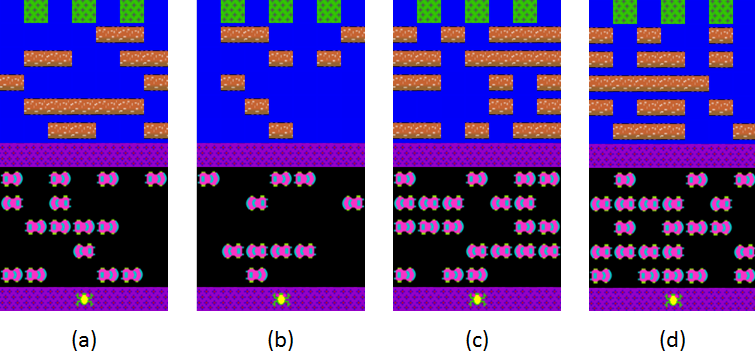}
\caption{(a) The Frogger map used for training. (b) The 25\% map used for testing. (c) The 50\% map used for testing. (d) The 75\% map used for testing.}
\label{fig:frogger_maps}
\end{figure*}
\begin{equation}
\label{eq:policyEquation}
Pr_{lc}(a)= \frac{e^{Pr_{l}(s,a,i)/\tau}}{\sum_{a'}e^{Pr_{l}(s,a',i)/\tau} }
\end{equation}
where $Pr_{l}(s,a,i)$ is the log probability of performing action $a$ in state $s$ according to the language-based critique policy using sequence $i$ as input. 

In the original policy shaping algorithm, the critique policy is constantly updated while the agent is learning. 
Since the language-based critique policy is trained offline, it does not have an opportunity to update itself, which can lead to an agent blindly following poor feedback. 
To address this, we make use of the $\tau$ parameter in Equation~\ref{eq:policyEquation} to control how much weight we place on the knowledge extracted from the language-based critique policy. 
In practice, we have found that the algorithm performs well when $\tau$ is initialized to be a small value that increases over the course of learning. 
This will cause the RL agent to begin learning by trusting the language-based critique policy and then shift towards relying on its own experience as time goes on. 
This allows the agent to disregard feedback that results in poor Q-values over time. 

Having done this, the RL agent now explores its environment as it normally would using policy shaping; however, the probability of the agent performing an action in a given state is defined as: 
\begin{equation}
Pr(a) = \frac{Pr_{q}(a)Pr_{lc}(a)}{\sum_{a'\in A}Pr_{q}(a')Pr_{lc}(a')}
\end{equation}

where the original probability obtained from human feedback, $Pr_{c}(a)$, is replaced with the probability obtained from the language-based critique policy. 

\section{Evaluation}
\label{sec:experiments}
To evaluate this technique, we examine how performs in training vitual agents to play the arcade game, Frogger. 
Specifically, we seek to show that using natural language to augment policy shaping enables reinforcement learning agents to speed up learning in unknown environments. 
In this section we will discuss our evaluation in which we compare agents trained with our technique against agents trained using a Q-learning algorithm as well as an agent trained using a baseline policy shaping algorithm with only access to behavior observations.

\subsection{Frogger}
\label{sec:frogger}
In these experiments we use the arcade game, Frogger (see Figure~\ref{fig:frogger_maps}), as a test domain. 
We chose Frogger because it is a discrete environment that can still be quite complex due to the specific mechanics of the environment. 
The learning agent's goal in this environment is to move from the bottom of the environment to the top while navigating the obstacles in the world. 
In this world, the obstacles move following a set pattern. 
Obstacles on on alternating rows will move on space to either the left or the right every time step. 
Moving outside of the bounds of the map, getting hit by a car, or falling into the water will result in the agent's death, which imposes a reward penalty of -10, reaching the goal earns the agent a reward of +100, and any other move taken in this environment will result in a small reward pentaly of -1. 
In this environment, the agent can take actions to move up, down, left, right, or choose to do nothing. 

We test our technique on three different Frogger environments. 
These environments, shown in Figures~\ref{fig:frogger_maps}~(b), (c), and (d), differ based on obstacle density. 
Specifically, we evaluate performance in maps in which spaces have a 25\% chance of containing an obstacle, a 50\% chance of containing an obstacle, and a 75\% chance of containing an obstacle. 

In addition, we evaluate agent performance in these environments under two different conditions: 1) a deterministic condition in which all actions execute normally, and 2) a stochastic condition in which actions have an 80\% chance of executing normally and a 20\% chance that the agent's action fails and it executes a different action instead. 

\subsection{Methodology}
In this section, we will discuss the experimental methodology used to both create and evaluate the language-based critique policy in Frogger. 
\subsubsection{Data Collection}

Since human teachers generate the natural language that is used for training, it is possible that mistakes will be made. 
Therefore, it is important to examine the effect that imperfect teachers will have on our technique. 
It is difficult to control for this type of error using actual humans, so for these experiments we use simulated human oracles to generate the required training observations and natural language.  
To create the behavior traces required for training, we trained $1000$ reinforcement learning agents with random starting positions to move one row forward while dodging obstacles on the training map in Figure~\ref{fig:frogger_maps}~(a). 
We chose this specific task to help eliminate any map-specific strategies that may be learned by using an agent trained to navigate the complete environment. 
To further help eliminate map-specific strategies, the states recorded for these training examples and then used in the remainder of these experiments encompass only a 3x3 grid surrounding the agent. 
This was done to help prevent the encoder-decoder network making spurious associations between the natural language annotations and potentially unrelated regions of the state space. 


Since we are using simulated humans to generate state and action traces, we use a grammar to create the natural language annotations that our system require. 
This grammar was constructing following the technique used in~\cite{harrison2017rationalization}.
This technique uses natural language utterances generated by humans to create a grammar in such a way that variances in human language are preserved and codified.
In order to produce a sentence, the grammar must first be provided with state and action information and then the grammar identifies the most appropriate grammar rule  to construct a natural language sentence.
The grammar is constructed such that each grammar rule can produce a large number of unique sentences.
Using this information, the grammar can then produce a natural language sentence describing the states and actions.

Since human trainers are likely to make mistakes, we test our technique's ability to deal with imperfect human trainers by introducing uncertainty into the natural language annotations generated by this grammar. 
Specifically, we create two different language-based critique models using the following simulated teachers: 1) teachers that use the correct grammar rule to provide feedback 80\% the time and and a random rule to provide feedback 20\% of the time (referred to from now on as the \textit{80\% training set}), and 2) teachers that use the correct grammar rule to provide feedback 60\% of the time and a random rule to provide feedback 40\% of the time (referred to from now on as the \textit{60\% training set}).
This will allow us to evaluate how robust our technique is to potential errors contained in the natural language feedback provided by human trainers. 
We feel that this also helps mitigate the regularity that is often present when using synthetic grammars.
It is important to note, that each of these training sets was generated using the same set of behavior traces. 
This makes the performance of the resulting critique policies directly comparable. 
After error was injected into each training set, duplicate training examples were removed.

To provide further evidence on the variability of the grammar with respect to these training sets, we also examined how often sentences repeated themselves in each training set. 
In the 80\% training set, the most seen sentence comprised 2.2\% of the total training set, which contained a total of 1433 training examples. 
On average, a sentence was repeated 0.19\% of the time. 
For the 60\% training set, the most seen sentence comprised 1.7\% of the training set of 1497 examples. 
On average, a sentence was repeated 0.17\% of the time. 
The disparity in training set size is due to duplicate training examples being removed after errors were introduced. 

\begin{figure*}
\centering
\includegraphics[scale=0.62]{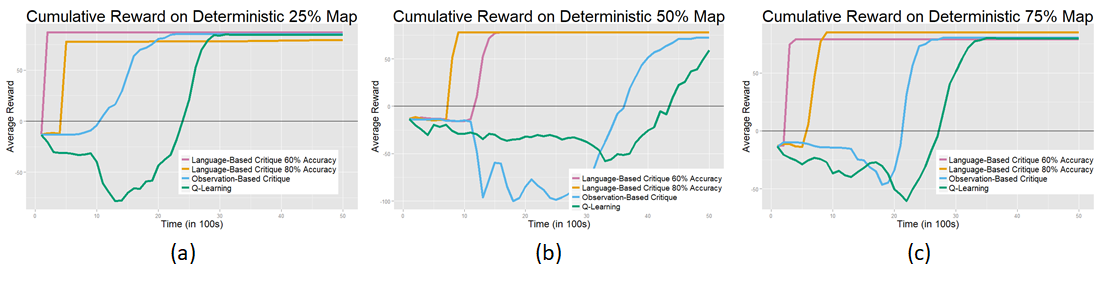}
\caption{Learning rates for agents on deterministic versions of (a) the 25\% map,(b) the 50\% map, and (c) the 75\% map. }
\label{fig:det-det_expt}
\end{figure*}

\begin{figure*}
\centering
\includegraphics[scale=0.62]{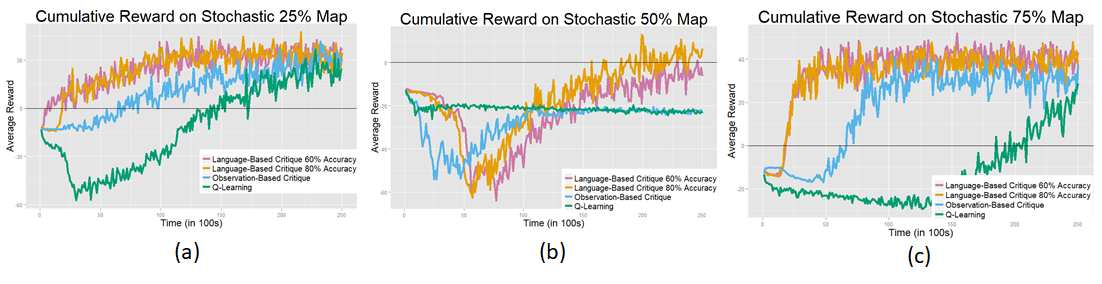}
\caption{Learning rates for agents on stochastic versions of the 25\% map (a), 50\% map (b), and 75\% map(c).}
\label{fig:det-stoch_expt}
\end{figure*}

\subsubsection{Training}
Using this dataset, we train an encoder-decoder network for 100 ephocs. 
Specifically we use an embedding encoder-decoder network with attention that is comprised two-layer recurrent neural networks composed of long short-term memory cells containing 300 hidden units each and an embedding size of 300. 
As mentioned previously, the network learns to translate between natural language descriptions and state and action information. 
For these experiments, the network learns to translate natural language generated by the grammar into the provided state and action information, which in this case is the 3x3 grid surrounding the agent as well as the action performed in that state. 

\subsubsection{Evaluation}
We ran experiments using four intelligent agents. The first is a baseline Q-learning agent with no access human feedback that we will refer to as the \textit{Q-learning} agent. The second is an agent trained using policy shaping that we will refer to as the \textit{observation-based critique} agent.
This agent has access to the state and action information that were used to train the language-based critique policy, which it uses as a positive action feedback signal. 
To help guard against poor examples in the training set, this agent also uses the parameter, $\tau$, to control how much weight is given to the action examples and to the agent's own experience. 
The final two agents are trained using our technique and we refer to them as the \textit{language-based critique 80\% accuracy} agent and the \textit{language-based critique 60\% accuracy} agent depending on which training set was used to generate the feedback oracle that the agent used during learning. 
Each of these agents was evaluated on each of the three unseen Frogger maps in the deterministic and stochastic conditions described previously. 
These test cases are meant to simulate how each agent performs in a simple (deterministic) environment as well as a more complex (stochastic) environment. 



For the deterministic test case, each agent was trained for 5000 episodes.
For the stochastic test case, each agent was trained for 25,000 episodes.
In all test cases, the learned policy was evaluated every 100 episodes and then the total cumulative reward earned during each episode was averaged over 100 total runs.

\subsection{Results}
The results for each agent on the deterministic Frogger maps can be seen in Figure~\ref{fig:det-det_expt} and the results for each agent on the stochastic Frogger maps can be seen in Figure~\ref{fig:det-stoch_expt}.
For both the language-based critique agent and the observation-based critique agent we tested several initial values and schedules for increasing the $\tau$ parameters. 
The graphs show the best results achieved in these experiments. 

As can be seen from Figure~\ref{fig:det-det_expt}, both language-based critique agents converge much faster than the Q-learning agent and the observation-based critique agent. 
Interestingly, the 60\% accurate language-based critique agent outperformed the 80\% accurate language-based critique agent on the 25\% map and the 75\% map.
It is important to note that the observation-based critique agent also consistently outperforms the Q-learning agent on each map, meaning that using observations still provides some benefit during training in unseen environments. 

Figure~\ref{fig:det-stoch_expt} shows the results that each agent obtained on the stochastic versions of the test Frogger maps.
For this set of experiments, both language-based critique agents outperform the other two agents on each map used for testing. 
Contrary to the performance in the deterministic environments, the 60\% accurate language-based critique agent and 80\% accurate language-based critique agent performed similarly in these environments.
It is also interesting to note that on the $50\%$ map, the language-based critique agents are the only ones to converge after the 25,000 training episodes. 
In the 25\% map and the 75\% map, the observation-based critique agent outperforms the Q-learning agent. 
Similar its performance in the deterministic environments, this shows that access to behavior observations still provides some amount of benefit in generalizing behavior to unseen environments.

\subsection{Discussion}
The first thing to note is that across all test cases the language-based critique agents either outperformed both the observation-based critique agent and the baseline Q-learning agent.
This shows that natural language provides knowledge useful for generalizing to unseen environments that cannot be obtained by simply looking at past observations.
In addition, this shows that our technique is robust to complexity in the learning environment as well as language error that may be present in human trainers. 
This is a significant result as this provides evidence that the encoder-decoder network can identify relevant features in the set of natural language annotations used for training even when the dataset contains a large amount of noise. 
These differences in performance were especially pronounced on the deterministic and stochastic 50\% maps, as well as the deterministic and stochastic 75\% maps.
In these cases, both language-based critique agents drastically outperformed both baselines.
The consistently positive results across both deterministic and stochastic environments further shows what a powerful tool language can be with respect to generalizing knowledge across many types of environments. 


One result that needs to be discussed is the performance of the 60\% accurate language-based critique agent with respect to the 80\% accurate language-based critique agent on the deterministic 25\% and 75\% maps. 
In each of these cases, the 60\% accurate agent outperformed the 80\% accurate agent, contrary to our intuition that the 80\% accurate agent should consistently outperform the 60\% accurate agent due to the additional error contained in the latter training set.
We hypothesize that this behavior can be explained due to model overfitting causing erratic behavior in these two cases. 
If the encoder-decoder network was overfitting part of the training set, then it is possible that the increased error introduced in the 60\% accurate agent had a regularizing effect on the network, which allowed it to better generalize to the states found in these two maps. 


\section{Limitations}
\label{sec:future-work}
While the results of our experiments provide strong evidence that our technique is effective at utilizing language to help learn generalizable knowledge, our technique is not without its limitations. 
First, our experiments made several simplifying assumptions that were necessary in order to control for the variance that accompanies human teachers. 
By using a grammar, we were able to control the amount of variance present in the annotations used for training. 
While we attempted to mitigate this by encoding our grammar with a large amount of variance and training the language-based critique model using training sets containing natural language error, naturally occurring language is likely to contain more variation than is present in our grammar. 

In addition, the natural language explanations used to train the language-based critique model were used to annotate single actions. 
Typically humans provide explanations of actions in context of a larger goal-based behavior trajectory and not on the level of individual actions. 
One way to improve this system is to enable it to learn from state/action explanations at varying levels of granularity. 

We also only explored how this technique can be applied in discrete environments. 
Using a discrete environment makes it easy to associate natural language annotations with state and action information. 
If this was done in a continuous environment then it would be much more difficult to determine what state or action should be associated with certain natural language annotations. 

Finally, we have only explored how this technique can be used to generalize to unseen environments within the same domain. 
It is unclear, however, if this technique could be used to aid in transfering knoweldge to agents learning similar tasks in different domains. 

\section{Conclusions}
\label{sec:conclusions}
Language is a powerful tool that humans use to generalize knowledge across a large number of states. 
In this work, we explore how language can be used to augment machine intelligence and give intelligent agents an expanded ability to generalize knowledge to unknown environments. 
Specifically, we show how neural machine translation techniques can be used to give action advice to reinforcement learning agents that generalizes across many different states, even if they have not been seen before. 
As our experiments have shown, this generalized model of advice enables reinforcement learning agents to quickly learn in unseen environments. 

In addition, this technique gives human teachers another way to train intelligent agents. 
The ability to augment human demonstration or critique with human feedback has the potential to significantly reduce the amount of effort required in order to train intelligent agents. 
This makes the task of training intelligent agents more approachable to potential human trainers.
It is even possible that this task could be crowdsourced in the future, drastically reducing the effort on the part of an individual trainer and making these types of agent training methods more appealing. 
Through this work, we hope to help bring down the language barrier that exists between humans and intelligent agents. 
By removing this barrier, we hope to enable humans to transfer more complex knowledge to intelligent agents, which should allow them to learn even more complex tasks in complex, unknown environments. 

\bibliography{bibliography}

\begin{thebibliography}{}

\bibitem[\protect\citeauthoryear{Argall \bgroup et al\mbox.\egroup
  }{2009}]{argall2009survey}
Argall, B.~D.; Chernova, S.; Veloso, M.; and Browning, B.
\newblock 2009.
\newblock A survey of robot learning from demonstration.
\newblock {\em Robotics and autonomous systems} 57(5):469--483.

\bibitem[\protect\citeauthoryear{Branavan \bgroup et al\mbox.\egroup
  }{2009}]{branavan2009reinforcement}
Branavan, S.~R.; Chen, H.; Zettlemoyer, L.~S.; and Barzilay, R.
\newblock 2009.
\newblock Reinforcement learning for mapping instructions to actions.
\newblock In {\em Proceedings of the Joint Conference of the 47th Annual
  Meeting of the ACL and the 4th International Joint Conference on Natural
  Language Processing of the AFNLP: Volume 1-Volume 1},  82--90.
\newblock Association for Computational Linguistics.

\bibitem[\protect\citeauthoryear{Branavan, Zettlemoyer, and
  Barzilay}{2010}]{branavan2010reading}
Branavan, S.; Zettlemoyer, L.~S.; and Barzilay, R.
\newblock 2010.
\newblock Reading between the lines: Learning to map high-level instructions to
  commands.
\newblock In {\em Proceedings of the 48th Annual Meeting of the Association for
  Computational Linguistics},  1268--1277.
\newblock Association for Computational Linguistics.

\bibitem[\protect\citeauthoryear{Cederborg \bgroup et al\mbox.\egroup
  }{2015}]{cederborg2015policy}
Cederborg, T.; Grover, I.; Isbell, C.~L.; and Thomaz, A.~L.
\newblock 2015.
\newblock Policy shaping with human teachers.
\newblock In {\em Twenty-Fourth International Joint Conference on Artificial
  Intelligence}.

\bibitem[\protect\citeauthoryear{Chernova and Thomaz}{2014}]{chernova2014robot}
Chernova, S., and Thomaz, A.~L.
\newblock 2014.
\newblock Robot learning from human teachers.
\newblock {\em Synthesis Lectures on Artificial Intelligence and Machine
  Learning} 8(3):1--121.

\bibitem[\protect\citeauthoryear{Griffith \bgroup et al\mbox.\egroup
  }{2013}]{griffith2013policy}
Griffith, S.; Subramanian, K.; Scholz, J.; Isbell, C.; and Thomaz, A.~L.
\newblock 2013.
\newblock Policy shaping: Integrating human feedback with reinforcement
  learning.
\newblock In {\em Advances in Neural Information Processing Systems},
  2625--2633.

\bibitem[\protect\citeauthoryear{Harrison, Ehsan, and
  Riedl}{2017}]{harrison2017rationalization}
Harrison, B.; Ehsan, U.; and Riedl, M.~O.
\newblock 2017.
\newblock Rationalization: A neural machine translation approach to generating
  natural language explanations.
\newblock {\em arXiv preprint arXiv:1702.07826}.

\bibitem[\protect\citeauthoryear{Krening \bgroup et al\mbox.\egroup
  }{2017}]{krening2017learning}
Krening, S.; Harrison, B.; Feigh, K.~M.; Isbell, C.~L.; Riedl, M.; and Thomaz,
  A.
\newblock 2017.
\newblock Learning from explanations using sentiment and advice in rl.
\newblock {\em IEEE Transactions on Cognitive and Developmental Systems}
  9(1):44--55.

\bibitem[\protect\citeauthoryear{Loftin \bgroup et al\mbox.\egroup
  }{2014}]{loftin2014learning}
Loftin, R.; Peng, B.; MacGlashan, J.; Littman, M.~L.; Taylor, M.~E.; Huang, J.;
  and Roberts, D.~L.
\newblock 2014.
\newblock Learning something from nothing: Leveraging implicit human feedback
  strategies.
\newblock In {\em Robot and Human Interactive Communication, 2014 RO-MAN: The
  23rd IEEE International Symposium on},  607--612.
\newblock IEEE.

\bibitem[\protect\citeauthoryear{Luong, Pham, and
  Manning}{2015}]{luong-pham-manning:2015:EMNLP}
Luong, M.-T.; Pham, H.; and Manning, C.~D.
\newblock 2015.
\newblock Effective approaches to attention-based neural machine translation.
\newblock In {\em Proceedings of the 2015 Conference on Empirical Methods in
  Natural Language Processing},  1412--1421.
\newblock Lisbon, Portugal: Association for Computational Linguistics.

\bibitem[\protect\citeauthoryear{MacGlashan \bgroup et al\mbox.\egroup
  }{2015}]{macglashan2015grounding}
MacGlashan, J.; Babes-Vroman, M.; desJardins, M.; Littman, M.~L.; Muresan, S.;
  Squire, S.; Tellex, S.; Arumugam, D.; and Yang, L.
\newblock 2015.
\newblock Grounding english commands to reward functions.
\newblock In {\em Robotics: Science and Systems}.

\bibitem[\protect\citeauthoryear{Matuszek \bgroup et al\mbox.\egroup
  }{2013}]{matuszek2013learning}
Matuszek, C.; Herbst, E.; Zettlemoyer, L.; and Fox, D.
\newblock 2013.
\newblock Learning to parse natural language commands to a robot control
  system.
\newblock In {\em Experimental Robotics},  403--415.
\newblock Springer.

\bibitem[\protect\citeauthoryear{Narasimhan, Kulkarni, and
  Barzilay}{2015}]{narasimhan2015language}
Narasimhan, K.; Kulkarni, T.~D.; and Barzilay, R.
\newblock 2015.
\newblock Language understanding for textbased games using deep reinforcement
  learning.
\newblock In {\em In Proceedings of the Conference on Empirical Methods in
  Natural Language Processing}.
\newblock Citeseer.

\bibitem[\protect\citeauthoryear{Sutton and Barto}{1998}]{suttonbarto1998}
Sutton, R.~S., and Barto, A.~G.
\newblock 1998.
\newblock {\em Reinforcement learning: An introduction}.
\newblock MIT Press.

\bibitem[\protect\citeauthoryear{Watkins and Dayan}{1992}]{watkins1992}
Watkins, C., and Dayan, P.
\newblock 1992.
\newblock Q-learning.
\newblock {\em Machine Learning} 8(3-4):279–292.

\bibitem[\protect\citeauthoryear{Watkins}{1989}]{watkins1989models}
Watkins, C.~J.
\newblock 1989.
\newblock {\em Models of delayed reinforcement learning}.
\newblock Ph.D. Dissertation, Ph. D. thesis, Cambridge University.

\end{thebibliography}
\bibliographystyle{aaai}

\end{document}